%% file: main.tex
\documentclass{article} 
\usepackage{iclr2026_conference,times}
\iclrfinalcopy

\input{math_commands.tex}

\usepackage{hyperref}
\usepackage{url}
\usepackage{float}
\usepackage{graphicx}
\usepackage{listings}
\usepackage{xcolor}
\usepackage{subcaption}

\usepackage{booktabs}
\usepackage{amsmath,amssymb}
\usepackage{listings}
\usepackage{float}
\usepackage{tikz}
\usetikzlibrary{arrows.meta,positioning,shapes.geometric}
\usepackage[ruled,vlined]{algorithm2e}
\usepackage{multirow}

\definecolor{mykeyword}{rgb}{0,0,0.7}   
\definecolor{mystring}{rgb}{0.58,0,0.82} 
\definecolor{mycomment}{rgb}{0,0.5,0}   
\definecolor{codebg}{rgb}{0.95,0.95,0.95} 

\title{MixtureKit: A General Framework for Composing, Training, and Visualizing Mixture-of-Experts Models}

\author{Ahmad Chamma\textsuperscript{1}$^\dagger$,
        Omar El Herraoui\textsuperscript{1},
        Guokan Shang\textsuperscript{1} \\
\textsuperscript{1}MBZUAI \\
 \small{
   $^\dagger$Correspondence: \texttt{ahmad.chamma@mbzuai.ac.ae}
 }
}

\begin{document}

\maketitle

\begin{abstract}
We introduce MixtureKit, a modular open-source framework for constructing, training, and analyzing Mixture-of-Experts (MoE) models from arbitrary pre-trained or fine-tuned models. MixtureKit currently supports three complementary methods: (i) \emph{Traditional MoE}, which uses a single router per transformer block to select experts, (ii) \emph{BTX} (Branch-Train-Mix), which introduces separate routers for each specified sub-layer enabling fine-grained token routing, and (iii) \emph{BTS} (Branch-Train-Stitch), which keeps experts fully intact and introduces trainable stitch layers for controlled information exchange between hub and experts. MixtureKit automatically modifies the model configuration, patches decoder and causal LM classes, and saves a unified checkpoint ready for inference or fine-tuning. We further provide a visualization interface to inspect per-token routing decisions, expert weight distributions, and layer-wise contributions. Experiments with multilingual code-switched data (e.g. Arabic-Latin) show that a BTX-based model trained using MixtureKit can outperform baseline dense models on multiple benchmarks. We release MixtureKit as a practical foundation for research and development of MoE-based systems across diverse domains.
The library is accessible at: \url{https://github.com/MBZUAI-Paris/MixtureKit}.
\end{abstract}

\section{Introduction}
\label{sec:intro}

Building Large Language Models (LLMs) is an area of growing interest for researchers and practitioners within the AI community.
This growth has been driven by the impressive performance of these models in a variety of downstream tasks \citep{survery_llm_powers} and has been accompanied by a rapid increase in the release of open-source models \citep{Casta_o_2023}.
These models cover diverse domains, such as healthcare \citep{sellergren2025medgemmatechnicalreport, Luo_2022}, software development \citep{rozière2024codellamaopenfoundation, hui2024qwen25codertechnicalreport},
and legal practice \citep{wu2023bloomberggptlargelanguagemodel, yang2023fingptopensourcefinanciallarge}.
To achieve higher performance and to expand the models' abilities in handling new tasks, previous work emphasized the need to scale up the models \citep{kaplan2020scalinglawsneurallanguage}, resulting in large dense models with up to trillions of parameters \citep{kimiteam2025kimik2openagentic, grattafiori2024llama3herdmodels}.
Despite the substantial value these models bring across domains and in users' daily tasks, they demand tremendous computational resources for optimal training and inference \citep{cottier2025risingcoststrainingfrontier}.
Furthermore, \citet{cf_luo} discovered that, during continual fine-tuning, catastrophic forgetting is more pronounced in larger models, likely due to their initial strong performance.
On the other hand,
\citet{haque2025catastrophicforgettingllmscomparative} reported that smaller models mitigate catastrophic forgetting and retain learning capacity, although they exhibit lower overall performance.

To leverage the advantages of both scales, it became imperative to introduce a new architecture combining large model capacity with the ability to update only a subset of parameters, reducing future forgetting, and maintaining a lower inference budget.
Mixture-of-Experts (MoE) models \citep{shazeer2017outrageouslylargeneuralnetworks} efficiently scale LLM capabilities by routing different inputs to specific sub-networks, allowing for fewer parameter updates.
With this new architecture, multiple models have been made available to the community that rank among the best published models, while offering outstanding performance and relatively low inference costs \citep{deepseekai2025deepseekv3technicalreport, openai2025gptoss120bgptoss20bmodel, yang2025qwen3technicalreport}.
However, the standard practice of pre-training MoE models from scratch incurs high computational costs.
Moreover, this approach does not provide any control over the specific domain of each expert, as it assumes that final convergence will occur after training without any prior knowledge of the domains \citep{sukhbaatar2024branchtrainmixmixingexpertllms}.
Osborne (2024) 
Furthermore, the work by \citet{Osborne_2024} highlighted that numerous models, including former state-of-the-art ones, remain unused after being surpassed by newer models.
Therefore, it is necessary to recycle outdated \textit{pre-trained} and \textit{fine-tuned} models into a unified state-of-the-art model, thus reducing computational costs and providing a strong initialization for further fine-tuning.

In this work, we introduce \texttt{MixtureKit}, an open-source python library for advanced Mixture-of-Experts architectures that helps accomplish this requested recycling.
Whereas some previous work has laid a solid ground for implementing certain architectures, it has primarily focused on well-known model families like Mistral, Llama, Phi, etc.
Accordingly, previous efforts have not provided a generalization beyond these families that could also accommodate custom-code models built on specialized architectures.
This package represents an additional effort to encourage the reuse of \textit{pre-trained} or \textit{fine-tuned} models, thereby substantially reducing computational training costs.
In addition, it automates the creation of the new MoE-based model, allowing users to concentrate on the subsequent stages of the training workflow without requiring extra skills.
Our main contributions are the following:
\begin{itemize}
    \item \textbf{Unified composer class}: The package provides a one-line merge function and a complete training pipeline, enabling full MoE training and usage without extensive low-level expertise. It automatically adjusts the model configuration, applies patches to the decoder and causal LM classes, and generates a unified checkpoint ready for inference or fine-tuning. High-level approach-specific helper functions are integrated into the new MoE model’s configuration, with necessary adjustments performed through regex pattern matching.
    \item \textbf{Advanced MoE strategies and load balancing}: The package supports advanced MoE schemes that focus on reusing HuggingFace checkpoints. These include \textit{Traditional MoE}, \textit{Branch-Train-miX} (BTX) and \textit{Branch-Train-Stitch} (BTS). The implementations, previously unavailable to the open community, are now accessible with full support for all model families, enabling further research and experimentation. In addition, it provides an implementation of the load-balancing principle among experts, dealing with the known inactive specialists issue in MoE models \citep{fedus2022switchtransformersscalingtrillion}.
    \item \textbf{Token routing visualization and statistics}: To ease interpretation and provide deeper insights into the model's internal decision-making process, the package includes a visualization tool that traces token routing paths at two levels: (1) High-level visualization, where each token is colored based on the expert that received the most votes across transformer blocks, revealing dominant routing choices, and (2) Low-level visualization, where each token is quantified with expert-specific percentage, reflecting its precise contribution to the corresponding output either per layer or on average.
\end{itemize}


\section{Related Work}

\paragraph{Mixture-of-Experts (MoE)}
The work of \citet{Jacobs1991Adaptive} introduced a new supervised learning paradigm for a system composed of multiple specialized neural networks, or experts coordinated by a gating network.
As the gating network assigns inputs to different experts, each expert network consequently learns to specialize in a particular subset of the input space.
They emphasized that this approach can achieve more efficient learning than a single dense network, especially for tasks subject to interference effects.
By confining weight updates to a subset of parameters, the system mitigates catastrophic forgetting and avoids the slow learning that often occurs in large single-network models when changes in one part of the network disrupt unrelated tasks.

\citet{shazeer2017outrageouslylargeneuralnetworks} revived the Mixture-of-Experts principle with their seminal paper.
The ability of the network to store and represent information is constrained by its number of parameters.
\textit{Conditional computation}, in which only specific parts of the network are activated for each input, has been theoretically proposed as a means of significantly increasing model capacity without corresponding increase in computational cost.
They proposed a \textit{Sparsely-Gated Mixture-of-Experts} (MoE) layer, where the gating network selects a sparse subset of $k$ experts to process each input, \textit{top-k} gate-routing mechanism.
The Switch Transformer \citep{fedus2022switchtransformersscalingtrillion} and GLaM \citep{du2022glamefficientscalinglanguage}, introduced by Google, were the pioneering models that demonstrated the effectiveness and scalability of the sparse MoE layers at the trillion-parameter level.
The former replaced the feed-forward network (FFN) layers with Switch layers containing multiple expert FFNs with a \textit{top-1} gating mechanism, in which each token was routed to a single expert for processing.
The study demonstrated that a Switch Transformer with $1.6$ trillion parameters exhibited higher efficiency than smaller dense models, such as the $11$ billion-parameter T5-XXL, when trained with equivalent computational resources.
GLaM employed a more sophisticated gating mechanism, whereby each token was directed to the \textit{top-2} experts from a total of 64 experts per MoE layer.
Nevertheless, these two models were not released as open-weight models for further experimentation by the community.

Mixtral \citep{jiang2024mixtralexperts} was the first open-weight model pre-trained from scratch using a mixture of $8$ experts, where $2$ are activated per input token.
It has been shown to match or outperform LLaMA 2 70B and GPT-3.5 in standard benchmarks. 
Its sparse design enables inference approximately six times faster than dense counterparts, improving computational efficiency and practicality for real-world deployment.

\paragraph{Model Merging}
The concept of model merging can be regarded as a related, albeit distinct, paradigm compared to traditional ensemble methods such as Random Forests \citep{chamma2023statisticallyvalidvariableimportance}.
It is evident that both approaches leverage the knowledge of multiple models; however, a distinction emerges in their execution.
Ensemble methods integrate the predictions of independently trained models during inference, thus improving robustness but resulting in a substantial computational burden.
In contrast, model merging techniques such as Model Soups \citep{wortsman2022modelsoupsaveragingweights} combine the parameters of fine-tuned models, often through simple averaging, into a single model that retains their collective strengths without incurring additional inference cost, marking a pivotal innovation that distills the advantages of ensembles into a deployable unified model.
Inspired by \textit{ensemble learning} and \textit{parameter averaging}, \citet{li2022branchtrainmergeembarrassinglyparalleltraining} introduced a new principle, \textit{Branch-Train-Merge} (BTM), in which a base model was replicated multiple times, each replica was trained on a separate subset of the original data, and the final output was obtained by combining the predictions of all replicas.
As for inference, they allow for either merging all experts into a single model to improve efficiency or selectively utilizing only the most relevant experts for a specific task to maximize performance.
Despite its strong performance, this architecture limited the ability to further fine-tune the individual experts' components within the unified structure.

\paragraph{MergeKit \citep{goddard2025arceesmergekittoolkitmerging}}
\texttt{MergeKit}, delivered by Acree-AI, is an open-source library focusing on providing the community with easy access to various merging techniques through a YAML syntax configuration file.
It enables the combination of model checkpoints, allowing their parameters to be merged to improve both performance and versatility, in response to the rapid growth of open-source language models.
It also provides Mixture-of-Experts (MoE) merging techniques, where the implemented gates can be initialized under: \textit{hidden} (based on the hidden state representations of the provided positive and negative prompts), \textit{cheap\_embed} (using only the raw token embeddings of the prompts) and \textit{random} (randomly initializing the MoE gates).
Nonetheless, the final model can only follow the direct MoE configuration of \texttt{Mixtral}, \texttt{DeepSeek} or \texttt{Qwen}, and therefore does not generalize.
Moreover, the package does not include open-source implementations for newly developed MoE merging techniques.

\paragraph{Mergoo}
\texttt{Mergoo}\footnote{https://github.com/Leeroo-AI/mergoo}, delivered by Leeroo-AI, was the first package to deliver open-source implementations of new MoE-based merging techniques such as BTX (section \ref{sec:btx}), along with simplified configuration and walkthrough examples.
However, this package focused mainly on hand-crafted model configuration files for known families such as Phi3, Mistral, and LLama, which limits its expansion to additional families.

\begin{figure}[!htb]
    \centering
    \includegraphics[width=\linewidth]{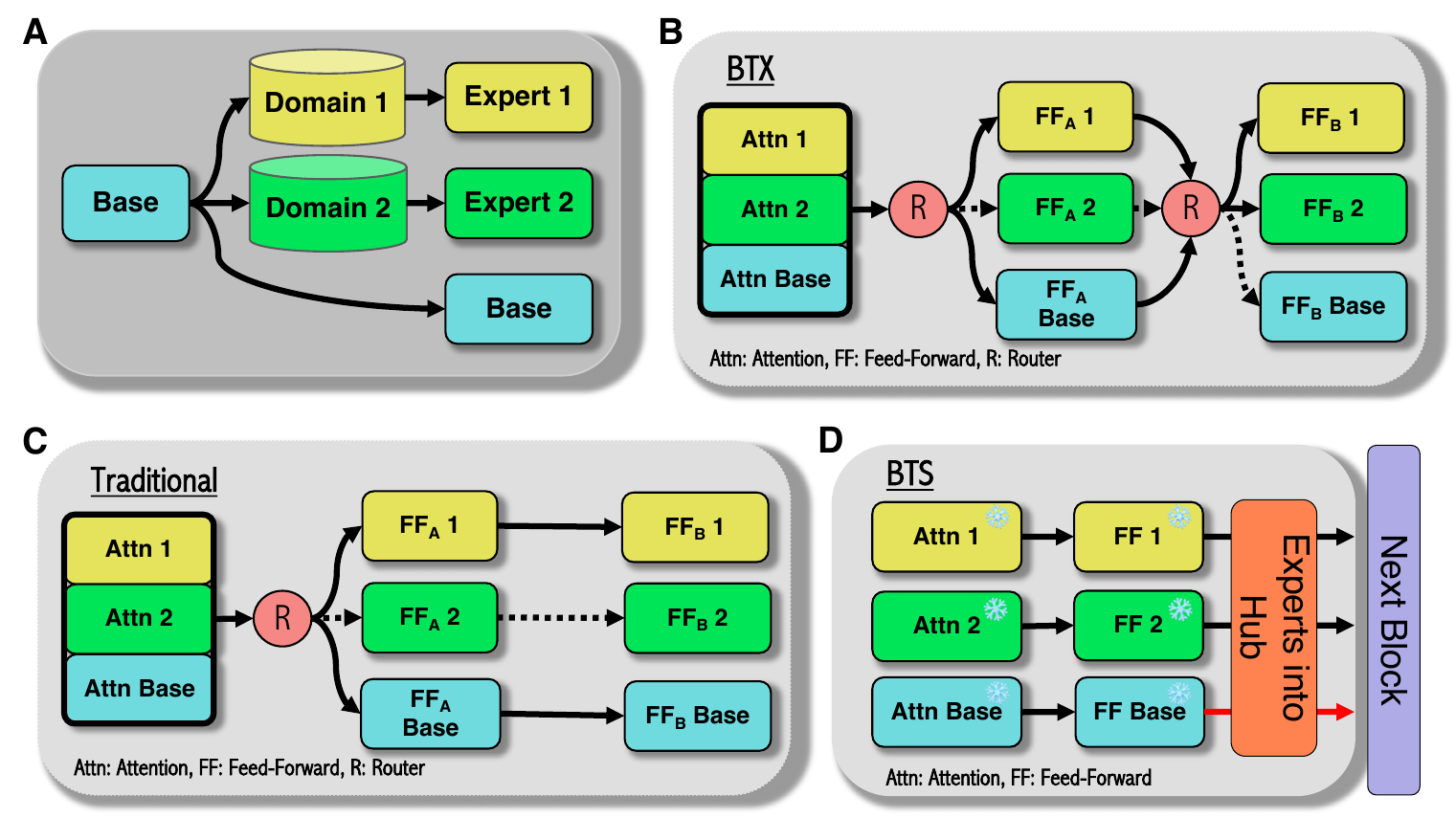}
    \caption{The workflow for building the new MoE-based unified model. (\textbf{A}) The common core component, where experts are prepared via continual pre-training or fine-tuning. This step encompasses the trained checkpoints hosted on the HuggingFace Hub \citep{wolf2020huggingfacestransformersstateoftheartnatural}. A replica of the base model is also included to transfer knowledge to the new model. Figures (\textbf{B--C}) show \textit{Branch-Train-miX (BTX)} and \textit{Traditional MoE} routing strategies respectively. The straight lines represent the experts selected under the top-2 routing scheme, while the dashed line indicates the optional additional expert that may be chosen depending on the configured value of top-$k$. $FF_{A}$ and $FF_{B}$ denote the internal projection components within the Feed-Forward networks, such as gate, up and down. (\textbf{D}) Steering-based strategies such as \textit{Branch-Train-Stitch (BTS)} where the input token is forwarded through both the frozen base and expert models. A trainable stitch layer is integrated to adjust the activations. The red line presents the final output derived from the base model after refinement with the experts.}
    \label{fig:moe-flow}
\end{figure}

\section{MixtureKit: Design and Implementation}
\label{sec:mixturekit}

\noindent\texttt{MixtureKit} provides a method-agnostic pipeline that composes a set of pretrained experts into a single checkpoint and patches the target architecture with either token routers (Traditional, BTX) or stitch layers (BTS). The system is driven entirely by a configuration dictionary (Fig.~\ref{fig:mixturekit-config}) and exposes a uniform interface that makes extending to new MoE variants a matter of registering a conversion rule for targeted submodules.

\subsection{Implemented Methods}
\paragraph{Branch-Train-miX (BTX)}
\label{sec:btx}
%
Whereas the previous direction was to pre-train an MoE model from scratch without any prior assignment of domain specialties to experts, a process that demanded tremendous computational resources for efficient training \citep{mu2025comprehensivesurveymixtureofexpertsalgorithms},  \citet{sukhbaatar2024branchtrainmixmixingexpertllms} introduced a recycling strategy called \textbf{Branch-Train-miX} (BTX) shown in Fig.~\ref{fig:moe-flow}-B.
Following the work of \citep{li2022branchtrainmergeembarrassinglyparalleltraining}, they started from the same premise where multiple instances of the selected base model were trained on different partitions of the original dataset, as illustrated in Fig.~\ref{fig:moe-flow}-A.
The feed-forward layers of these models are reconfigured as individual experts within a new MoE layer, while a trainable routing network directs each token to the most appropriate expert(s) according to the chosen mechanism.
Unless otherwise specified, layers such as attention and embeddings are merged through parameter averaging across the instances, yielding a unified backbone.
The unified MoE model can, in principle, be adapted using two fine-tuning strategies.
An option is to update only the integrated routers while keeping all other parameters frozen, which offers computational efficiency but results in a slight degradation of performance; alternatively, full fine-tuning of all model components can be employed, particularly after parameter averaging, which is more computationally demanding but necessary to achieve optimal accuracy and fully realize the potential of the composed model.
Despite promising results, this method relies on decomposing the components of different experts and reintegrating them according to the MoE framework; as a result, the merged model does not support the flexibility and interpretability offered by a modular structure, in which experts are kept separate and intact.

\paragraph{Traditional MoE}
As shown in Fig.~\ref{fig:moe-flow}-B, \emph{BTX} introduces independent routers at each internal projection within the FFN (gate, up, and down projections) which provides a unified model with greater degrees of freedom in selecting a specific path among experts. However, this comes at the cost of increased routing complexity. 
To align with the architecture presented in the work by
\citet{jiang2024mixtralexperts}, we propose a variant of \textit{BTX} called \textbf{Traditional MoE} as presented in Fig.~\ref{fig:moe-flow}-C. which employs routing through entire FFN blocks.
For every token, the router chooses one expert (or top-$K$ experts), and the selected experts are then used consistently across all internal projections.
The unified model can be fine-tuned in the same way as BTX, but with fewer parameters since there are fewer routers.
Although inputs are routed through the different experts, some of them are unlikely to be selected for processing, leading to the phenomenon of \textit{dead experts}.
The key to bypassing this issue is the addition of a loss, referred to as the load balancing loss, with a hyperparameter $\alpha$, which encourages the model to distribute workload more evenly among experts.
However, a trade-off is needed in selecting $\alpha$ to prevent both significant degradation in the model’s performance and instability during training.

\paragraph{Branch-Train-Stitch (BTS)}
%
Although router-based models paved the way for the construction of recycled MoE models inspired by ensemble learning and parameter averaging, they lack a plug-in format that would allow for the seamless integration of new experts specialized in additional domains.
To produce a unified model that is more inherently interpretable, \citet{zhang2025btsharmonizingspecializedexperts} described a new flexible method called \textbf{Branch-Train-Stitch} (BTS).
As presented in Fig.~\ref{fig:moe-flow}-D, once all experts (i.e. replicas) have been trained on the respective subsets of the original data, no parameter averaging is applied in this method; instead, each expert is retained as-is in the final model structure.
To interchange information between experts and the hub (i.e. base) model, a special trainable bidirectional \textit{StitchLayer} is added on top of selected transformer blocks, including the final block, at a predefined frequency to steer the corresponding activations.
This layer consists of linear projections and a linear gate, with only one direction active at a time, while all the parameters of the underlying transformer blocks remain frozen.
The first direction, \texttt{experts-into-hub}, refines the activations of the target transformer block.
It linearly projects the hidden states of the experts in the hidden representation space of the hub model and computes the contributions of each expert, including the hub itself, to the output.
This is done by first applying a linear gate ($w_{gate}$) to the hub’s hidden state ($h_0$), followed by a softmax across all experts and the hub to obtain their relative contributions.
The second direction, \texttt{hub-into-experts}, adjusts the activations of the experts by first linearly projecting the hub’s hidden state into the hidden representation space of the experts.
Instead of using a softmax function, this direction applies a sigmoid function to determine the relative contribution of the hub and each expert’s previous hidden state to the updated states.
BTS enables the merging of specialized experts in a flexible plug-in format, where training focuses mainly on the stitch lightweight layers, making it efficient and requiring low computational cost.
However, one drawback of this approach is the memory challenge that may arise when fitting all the full experts

\subsection{End-to-End Workflow}

\begin{figure}[!htb]
\centering
\begin{minipage}{\linewidth}
\begin{lstlisting}
config = {
  "moe_method": "btx",
  "stitch_freq": 5,
  "model_type": "new_model_type",
  "num_experts_per_tok": 2,
  "experts": [
    {"expert_name": "base_expert", "model_id": "expert_base_checkpoint"},
    {"expert_name": "expert_1",    "model_id": "expert_1_checkpoint"},
    {"expert_name": "expert_2",    "model_id": "expert_2_checkpoint"},
  ],
  "router_layers": ["mlp.gate_proj", "mlp.up_proj", "mlp.down_proj"],
  "alpha": 0, "router_layers_index": []
}
\end{lstlisting}
\end{minipage}
\caption{Example of configuration dictionary \texttt{config} merging the feed-forward layers of three models under the Branch-Train-miX strategy (BTX) with two experts activated and load balancing disabled.}
\label{fig:mixturekit-config}
\end{figure}

%

\paragraph{User-Centric Configuration.}  
The entire build process is controlled by a single dictionary configuration object, shown in Fig.~\ref{fig:mixturekit-config}.
The key fields directly specify the design choices of the unified MoE model.
The \texttt{moe\_method} selects the integration strategy (e.g., \texttt{btx},  \texttt{traditional} or \texttt{bts}), while \texttt{model\_type} defines the identifier of the unified checkpoint.
The list of \texttt{experts} provides the Hugging Face model IDs for the base model and the domain-specialized checkpoints for merging.
Sparsity at inference time is controlled by \texttt{num\_experts\_per\_tok}, which determines how many experts are activated for each token.
This setting is equivalent to the top-$k$ parameter in routing, where only the $k$ highest-scoring experts are selected to contribute to the final output.
The fields \texttt{router\_layers} and \texttt{router\_layers\_index} specify which submodules (e.g., \texttt{mlp} for MoEs or \texttt{attn} for MoAs \citep{fu2024moamixturesparseattention}) and which indexed transformer blocks are converted into MoE form, respectively.
Additional parameters such as \texttt{stitch\_freq} (for stitch-based methods) and \texttt{alpha} (for load-balancing regularization) further control the training behavior.
Once defined, a single call to \texttt{build\_moe(config)} produces a fully functional, \texttt{transformers}-compatible checkpoint.
This makes experiments repeatable and easily shareable without requiring modifications to the model code.


\paragraph{Expert Composition.}
The process begins with the \texttt{compose()} method, where each expert model is loaded and its parameters are iteratively integrated into a unified state dictionary.
For each parameter, the system determines whether it should be shared between experts or assigned individually to an expert-specific namespace.
The shared parameters are averaged across all experts with shape-aware alignment, ensuring compatibility even when the hidden dimensions differ slightly.
Parameters designated for expert-specific conversion are stored under structured namespaces (e.g., \texttt{experts.expert\_i.weight}), producing a coherent MoE parameter layout that remains fully compatible with the \texttt{transformers} library.
%

%


\paragraph{Architecture Patching and Method Specialization.}  
After expert composition, \texttt{MixtureKit} rewrites the destination model’s architecture to host the MoE extensions. This process is orchestrated in the \texttt{save\_checkpoint()} function, which copies the original \texttt{modeling\_<base>.py} and \texttt{configuration\_<base>.py} files, renames them to match the new \texttt{model\_type}, and updates imports and class names via \texttt{\_replace\_type\_dependency}.
Targeted edits then inject the MoE logic: \texttt{\_replace\_script} rewrites \texttt{nn.Linear} and \texttt{Conv1D} layers inside the \texttt{Attention} and \texttt{MLP} classes, while \texttt{\_modify\_decoder} and \texttt{\_modify\_model} adjust the decoder and model classes to incorporate gating and load-balancing losses.  

For router-based methods (\emph{Traditional}, \emph{BTX}), the patched classes are used to replace the original linear layers with MoE-aware modules (\texttt{convert\_linear\_to\_moe}), enabling a gating network that selects the top-$k$ experts at each step.
In contrast, stitch-based methods (\emph{BTS}) do not modify linear layers; instead, \texttt{\_patch\_stitches} augments the model’s forward path by introducing parallel expert streams and inserting \texttt{StitchLayer} modules that blend hub and expert activations at configurable depths controlled by the \texttt{stitch\_freq} parameter in the user configuration (Fig.~\ref{fig:mixturekit-config}).
Importantly, these specializations change only the forward computation and parameter organization, while the composition and saving pipeline remain identical.
This modularity ensures that the framework is naturally extensible to future MoE strategies with minimal additional code.

\paragraph{Extensibility.}  
Adding a new MoE variant only requires implementing a small adapter module that provides a conversion function for the chosen layers or a stitch integration function for hub–expert fusion. Once registered in the configuration, the new method automatically benefits from \texttt{MixtureKit}'s loading, saving, and compatibility pipeline, without touching the core composer logic.

\subsection{Visualization and Interpretability}
\label{sec:visualization}

\begin{figure}[!htb]
    \centering
    \includegraphics[width=\linewidth]{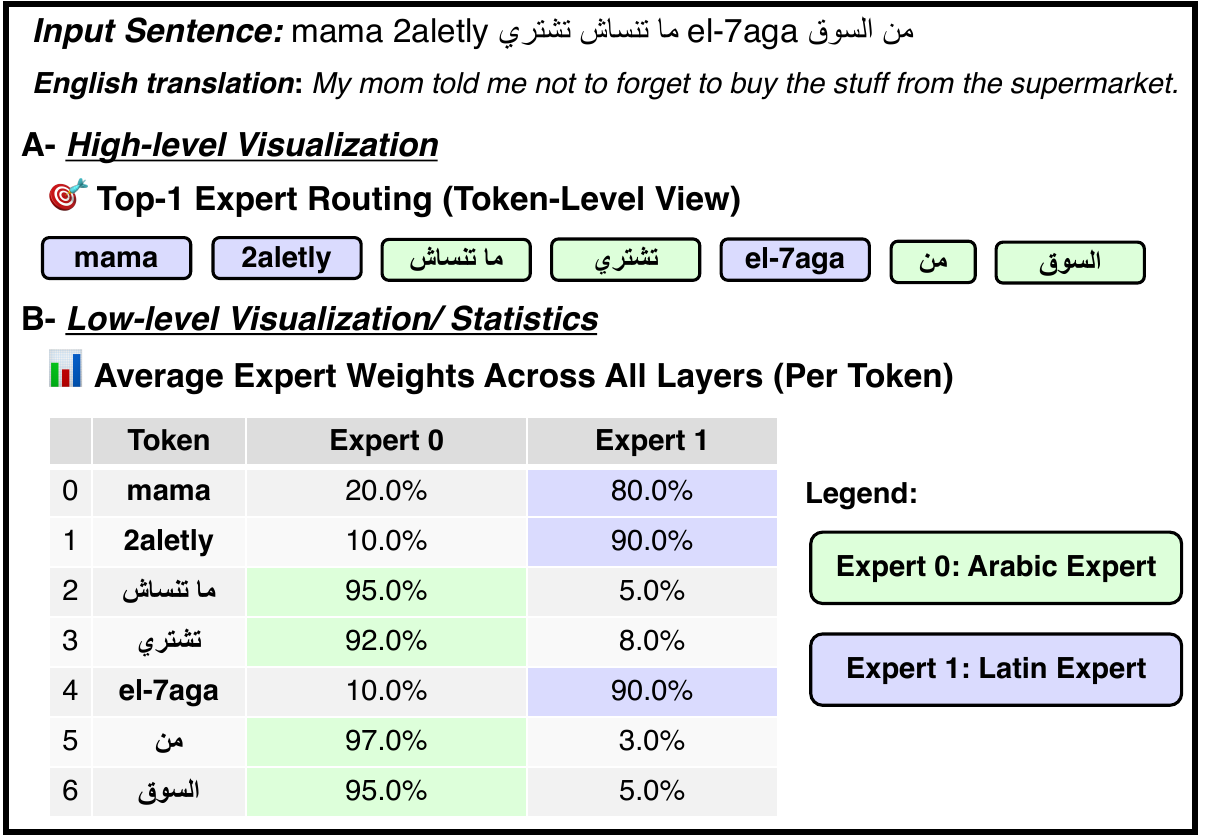}
    \caption{MixtureKit Token-Routing Visualizer, showing the main interface with the input prompt: (\textbf{A}) High-level color-coded token assignments showing expert specialization at a glance, and (\textbf{B}) Average expert weights across all selected layers for each token.}
    \label{fig:moe-viz}
\end{figure}

An important aspect of \texttt{MixtureKit} is its visualization interface, which provides researchers with real-time information on token routing decisions. The tool is implemented as an interactive \textit{Streamlit} application and supports all MoE configurations that rely on token-level routing, namely the \textit{Traditional} and \textit{BTX} variants.

\paragraph{Routing Mechanism.}
At the core of both Traditional and BTX routing is a gating function that maps each token representation to expert scores. 
Let $h_{t}^{(\ell,p)} \in \mathbb{R}^{d}$ denote the hidden state of token $t$ at transformer block $\ell$ and internal projection $p \in \{\text{gate}, \text{up}, \text{down}\}$. 
Each projection is paired with a gating matrix $W^{(\ell,p)} \in \mathbb{R}^{d \times E}$ that produces unnormalized logits with $E$ denoting the number of experts:
\[
g_{t,e}^{(\ell,p)} = h_{t}^{(\ell,p)} W^{(\ell,p)}, \qquad e \in \{1,\ldots,E\}.
\]
The router then selects the top-$k$ experts
\[
S_{t}^{(\ell,p)} = \text{TopK}\!\bigl(g_{t}^{(\ell,p)}, k\bigr),
\]
and assigns normalized weights using a softmax function
\[
w_{t,e}^{(\ell,p)} = \frac{\exp(g_{t,e}^{(\ell,p)})}{\sum_{j \in S_{t}^{(\ell,p)}} \exp(g_{t,j}^{(\ell,p)})}, \qquad e \in S_{t}^{(\ell,p)}.
\]

In \emph{BTX}, each projection (gate, up, down) has its own router $W^{(\ell,p)}$, so a token may follow different experts' components within the same FFN block. 
In \emph{Traditional MoE}, a single router $W^{(\ell)}$ is shared across all three projections, ensuring a consistent expert choice throughout the block.

To provide an interpretable perspective on routing, MixtureKit's visualization aggregates and displays expert normalized weights across the full depth of the model. 
For token $t$, the aggregated contribution of expert $e$ is
\[
\overline{w}_{t,e} = \frac{1}{|\mathcal{L}_{t}|} \sum_{(\ell,p) \in \mathcal{L}_{t}} w_{t,e}^{(\ell,p)}, 
\qquad \mathcal{L}_{t} = \{(\ell,p) : t \text{ is processed at block } \ell, \text{ projection } p\}.
\]

Figure~\ref{fig:moe-viz} illustrates both perspectives: 
(A) high-level token assignments based on the dominant expert $\arg\max_{e} \overline{w}_{t,e}$, and 
(B) detailed statistics showing the aggregated weights $\overline{w}_{t,e}$ throughout all transformer blocks and projections.
This combination enables both intuitive inspection of token-level specialization and fine-grained analysis of how expert usage evolves across the model depth.
In addition, the MixtureKit visualization interface allows users to restrict the aggregation to specific projections (e.g., only gate or up layers) at specific transformer blocks, facilitating a more detailed examination of how expert choices evolve throughout the model.

This tool has proven valuable for diagnosing expert under-utilization, detecting routing collapse (where one expert receives most tokens), and studying expert specialization in code-switched or multilingual scenarios.
It can be applied to any method where the contribution of each expert is easily accessible within the model.

\section{Practical Example: Sript-specialized experts}
Egyptian Arabic, also referred to as \textit{Masri}, is the most prevalent Arabic dialect, with a population of over $100$ million native speakers in Egypt and a high degree of mutual intelligibility throughout the Arab world.
It exhibits notable disparities from \textit{Modern Standard Arabic} (MSA) with respect to phonology, vocabulary, and grammar.
A distinctive aspect of this dialect is its frequent use of two writing systems. Egyptian Arabic speakers often employ both the traditional Arabic script and a Latin-based script, commonly known as Arabizi or Franco-Arabic \citep{egyptian_dialect}.
Multi-script Egyptian Arabic can be addressed by integrating script-specific experts, one trained on Arabic-script data and another on Latin-script (Arabizi) data, into a unified Mixture-of-Experts (MoE) model that dynamically routes tokens to the appropriate expert.
This modular design facilitates scalable adaptation across scripts while preserving high performance and efficiency.

To demonstrate the potential of the proposed library, we adopt the experimental setup from \citep{shang2025nilechategyptianlanguagemodels} and try to reproduce the results using \texttt{MixtureKit}.
To begin with, the base model \texttt{Gemma3-4B-pt} \footnote{https://huggingface.co/google/gemma-3-4b-pt} was continually pre-trained separately on available Arabic and Latin script datasets to develop script-specific experts.
Secondly, the pre-trained experts were integrated with the base model employing the \textit{BTX}, leading to a novel MoE model comprising three experts, two of which are active per input, with a total of 6B activated parameters.
This configuration defines the announced \textit{Nile-Chat-3x4B-A6B model} where incorporating the base model as an additional expert provided broader general knowledge and enhanced English capabilities, extending beyond the script-specialized experts.
For comparison, the two script-specialized experts were also merged independently of the base model, resulting in the outlined \textit{Nile-Chat-2x4B-A6B} variant.
The unified MoE models were trained in two stages.
Initially, Supervised Fine-Tuning (SFT) was performed employing a Low Rank Adaptation (LoRA) configuration with an alpha value set to $512$, a learning rate of $1e-4$, and an effective batch size of $256$.
In order to maintain the efficacy of the English-centric base model, which functions as a third expert, a proportion of English instructions have been incorporated from Wild-Chat \footnote{https://huggingface.co/datasets/allenai/WildChat-1M}.
Secondly, Direct Preference Optimization (DPO) has been implemented as the final alignment stage.

\begin{table}[!htb]
\centering
\renewcommand{\arraystretch}{1.13}
\setlength{\tabcolsep}{4pt}
\resizebox{\columnwidth}{!}{
\begin{tabular}{p{4cm}ccccc}
\toprule
\textbf{\small Model} &
\shortstack{\textbf{\small Average} \\ \textbf{\small Arabic}} &
\shortstack{\textbf{\small Average} \\ \textbf{\small Latin}} &
\shortstack{\textbf{\small Translation} \\ \textbf{\small Long (chrf)}} &
\shortstack{\textbf{\small Translation} \\ \textbf{\small Short (chrf)}} &
\shortstack{\textbf{\small Transliteration} \\ \textbf{\small (chrf)}} \\
\midrule   
\small \textbf{Nile-Chat-4B-Arabic-Expert}                    & 53.21 & 44.63 &
58.81 & 52.7 & 26.21 \\
\small \textbf{Nile-Chat-4B-Latin-Expert}                    & 48.85 & 48.06 &
37.09 & 31.27 & 80.59 \\
\midrule   
\small \textbf{Nile-Chat-4B}                    & 53.01 & 49.07 & 58.4 & 52.01 & 80.44 \\
\small \textbf{Nile-Chat-2x4B-A6B}              & \underline{55.87} & \underline{52.32} & \underline{61.59} & \underline{53.71} & \underline{83.89} \\
\small \textbf{Nile-Chat-3x4B-A6B}              & 55.74 & 51.23 & \textbf{61.9} & \textbf{55.37} & \textbf{83.97} \\
\small \textbf{Nile-Chat-12B}                   & \textbf{60.0} & \textbf{52.61} & 60.61 & 53.53 & 80.97 \\
\bottomrule
\end{tabular}}
\caption{Performance comparison of Arabic/Latin experts, Nile-Chat dense models (4B and 12B) and \textit{BTX}-based counterparts across Arabic, Latin and generation benchmarks.
The highest scores are indicated in \textbf{bold}, the second-highest are \underline{underlined}.}
\label{tab:results}
\end{table}

To evaluate the performance of the different experts and models, we followed the evaluation procedure described in previous work \citep{shang2025nilechategyptianlanguagemodels}, focusing on the average of all Egyptian bench-
marks in both Arabic and Latin scripts (measured with \textit{accuracy} or \textit{normalized accuracy}), as well as the generation benchmarks related to translation and transliteration (measure with \textit{chrF}).
As illustrated in Table~\ref{tab:results}, we successfully reproduced the findings, showing that \textbf{Nile-Chat-3x4B-A6B} and \textbf{Nile-Chat-2x4B-A6B} offer a compromise between the dense 4B and 12B models for discriminative tasks in Arabic script (53.01 $<$ \underline{55.87} $<$ \textbf{60.0}), perform comparably on Latin script (\underline{52.32} $\approx$ \textbf{52.61}), and outperform them in tasks requiring extensive text generation, achieving the top-2 highest scores across all translation and transliteration tasks and metrics.

\section{Conclusion}  
In this paper, we introduced \texttt{MixtureKit}, providing a practical step forward to make Mixture-of-Experts (MoE) research widely accessible.
By allowing the reuse of existing \textit{pre-trained} or {fine-tuned} checkpoints, systematically composing them into unified models, and adapting architectures for various routing and stitching strategies, it reduces the need to train costly MoE models from scratch.
Through its visualization interface, \texttt{MixtureKit} also makes expert behavior interpretable, helping diagnose routing imbalance and study specialization dynamics in multilingual or domain-specific settings.
Beyond the strong empirical results we report, the modular design of the framework reduces the barrier to experimentation with new MoE variants, offering the community a flexible foundation for future research and development.
The provision of an open-source toolkit for the merging of model checkpoints is intended to promote collaboration among researchers, developers, and practitioners worldwide, thereby fostering innovation and knowledge sharing.

\section{Future Work}
Since the library modifies submodules with MoE components based on regex pattern matching, it relies on the HuggingFace versioning of models' configuration files.
We aim to support as many versions as possible.
In addition, the new configuration files currently lack the optimizations introduced in recent inference packages, such as the recently introduced MoE kernels in \textit{vLLM} \footnote{https://github.com/vllm-project/vllm}.
MixtureKit is a dynamic project dedicated to continuously incorporating new methodologies through collaboration with the open-source community.
Additionally, while most of the literature focuses on merging models with the same architecture, an interesting direction is to explore cross-architecture merging.

\bibliography{iclr2026_conference}
\bibliographystyle{iclr2026_conference}

\appendix
\section{LLM Usage}
We use LLMs solely to polish writing and clarify ideas, keeping all scientific reasoning human-driven.
The model acts only as a stylistic assistant, enhancing readability without contributing content.

\end{document}

%% file: math_commands.tex
\usepackage{amsmath,amsfonts,bm}

\def\eqref#1{equation~\ref{#1}}

\def\1{\bm{1}}

\DeclareMathAlphabet{\mathsfit}{\encodingdefault}{\sfdefault}{m}{sl}
\SetMathAlphabet{\mathsfit}{bold}{\encodingdefault}{\sfdefault}{bx}{n}



%% file: iclr2026_conference.bib
@misc{goddard2025arceesmergekittoolkitmerging,
      title={Arcee's MergeKit: A Toolkit for Merging Large Language Models}, 
      author={Charles Goddard and Shamane Siriwardhana and Malikeh Ehghaghi and Luke Meyers and Vlad Karpukhin and Brian Benedict and Mark McQuade and Jacob Solawetz},
      year={2025},
      eprint={2403.13257},
      archivePrefix={arXiv},
      primaryClass={cs.CL},
      url={https://arxiv.org/abs/2403.13257}, 
}

@misc{sukhbaatar2024branchtrainmixmixingexpertllms,
      title={Branch-Train-MiX: Mixing Expert LLMs into a Mixture-of-Experts LLM}, 
      author={Sainbayar Sukhbaatar and Olga Golovneva and Vasu Sharma and Hu Xu and Xi Victoria Lin and Baptiste Rozière and Jacob Kahn and Daniel Li and Wen-tau Yih and Jason Weston and Xian Li},
      year={2024},
      eprint={2403.07816},
      archivePrefix={arXiv},
      primaryClass={cs.CL},
      url={https://arxiv.org/abs/2403.07816}, 
}

@misc{zhang2025btsharmonizingspecializedexperts,
      title={BTS: Harmonizing Specialized Experts into a Generalist LLM}, 
      author={Qizhen Zhang and Prajjwal Bhargava and Chloe Bi and Chris X. Cai and Jakob Foerster and Jeremy Fu and Punit Singh Koura and Ruan Silva and Sheng Shen and Emily Dinan and Suchin Gururangan and Mike Lewis},
      year={2025},
      eprint={2502.00075},
      archivePrefix={arXiv},
      primaryClass={cs.CL},
      url={https://arxiv.org/abs/2502.00075}, 
}

@misc{shang2025nilechategyptianlanguagemodels,
      title={Nile-Chat: Egyptian Language Models for Arabic and Latin Scripts}, 
      author={Guokan Shang and Hadi Abdine and Ahmad Chamma and Amr Mohamed and Mohamed Anwar and Abdelaziz Bounhar and Omar El Herraoui and Preslav Nakov and Michalis Vazirgiannis and Eric Xing},
      year={2025},
      eprint={2507.04569},
      archivePrefix={arXiv},
      primaryClass={cs.CL},
      url={https://arxiv.org/abs/2507.04569}, 
}

@misc{jiang2024mixtralexperts,
      title={Mixtral of Experts}, 
      author={Albert Q. Jiang and Alexandre Sablayrolles and Antoine Roux and Arthur Mensch and Blanche Savary and Chris Bamford and Devendra Singh Chaplot and Diego de las Casas and Emma Bou Hanna and Florian Bressand and Gianna Lengyel and Guillaume Bour and Guillaume Lample and Lélio Renard Lavaud and Lucile Saulnier and Marie-Anne Lachaux and Pierre Stock and Sandeep Subramanian and Sophia Yang and Szymon Antoniak and Teven Le Scao and Théophile Gervet and Thibaut Lavril and Thomas Wang and Timothée Lacroix and William El Sayed},
      year={2024},
      eprint={2401.04088},
      archivePrefix={arXiv},
}

@misc{sellergren2025medgemmatechnicalreport,
      title={MedGemma Technical Report}, 
      author={Andrew Sellergren and Sahar Kazemzadeh and Tiam Jaroensri and Atilla Kiraly and Madeleine Traverse and Timo Kohlberger and Shawn Xu and Fayaz Jamil and Cían Hughes and Charles Lau and Justin Chen and Fereshteh Mahvar and Liron Yatziv and Tiffany Chen and Bram Sterling and Stefanie Anna Baby and Susanna Maria Baby and Jeremy Lai and Samuel Schmidgall and Lu Yang and Kejia Chen and Per Bjornsson and Shashir Reddy and Ryan Brush and Kenneth Philbrick and Mercy Asiedu and Ines Mezerreg and Howard Hu and Howard Yang and Richa Tiwari and Sunny Jansen and Preeti Singh and Yun Liu and Shekoofeh Azizi and Aishwarya Kamath and Johan Ferret and Shreya Pathak and Nino Vieillard and Ramona Merhej and Sarah Perrin and Tatiana Matejovicova and Alexandre Ramé and Morgane Riviere and Louis Rouillard and Thomas Mesnard and Geoffrey Cideron and Jean-bastien Grill and Sabela Ramos and Edouard Yvinec and Michelle Casbon and Elena Buchatskaya and Jean-Baptiste Alayrac and Dmitry Lepikhin and Vlad Feinberg and Sebastian Borgeaud and Alek Andreev and Cassidy Hardin and Robert Dadashi and Léonard Hussenot and Armand Joulin and Olivier Bachem and Yossi Matias and Katherine Chou and Avinatan Hassidim and Kavi Goel and Clement Farabet and Joelle Barral and Tris Warkentin and Jonathon Shlens and David Fleet and Victor Cotruta and Omar Sanseviero and Gus Martins and Phoebe Kirk and Anand Rao and Shravya Shetty and David F. Steiner and Can Kirmizibayrak and Rory Pilgrim and Daniel Golden and Lin Yang},
      year={2025},
      eprint={2507.05201},
      archivePrefix={arXiv},
      primaryClass={cs.AI},
      url={https://arxiv.org/abs/2507.05201}, 
}

@misc{wolf2020huggingfacestransformersstateoftheartnatural,
      title={HuggingFace's Transformers: State-of-the-art Natural Language Processing}, 
      author={Thomas Wolf and Lysandre Debut and Victor Sanh and Julien Chaumond and Clement Delangue and Anthony Moi and Pierric Cistac and Tim Rault and Rémi Louf and Morgan Funtowicz and Joe Davison and Sam Shleifer and Patrick von Platen and Clara Ma and Yacine Jernite and Julien Plu and Canwen Xu and Teven Le Scao and Sylvain Gugger and Mariama Drame and Quentin Lhoest and Alexander M. Rush},
      year={2020},
      eprint={1910.03771},
      archivePrefix={arXiv},
      primaryClass={cs.CL},
      url={https://arxiv.org/abs/1910.03771}, 
}

@misc{openai2025gptoss120bgptoss20bmodel,
      title={gpt-oss-120b \& gpt-oss-20b Model Card}, 
      author={OpenAI and : and Sandhini Agarwal and Lama Ahmad and Jason Ai and Sam Altman and Andy Applebaum and Edwin Arbus and Rahul K. Arora and Yu Bai and Bowen Baker and Haiming Bao and Boaz Barak and Ally Bennett and Tyler Bertao and Nivedita Brett and Eugene Brevdo and Greg Brockman and Sebastien Bubeck and Che Chang and Kai Chen and Mark Chen and Enoch Cheung and Aidan Clark and Dan Cook and Marat Dukhan and Casey Dvorak and Kevin Fives and Vlad Fomenko and Timur Garipov and Kristian Georgiev and Mia Glaese and Tarun Gogineni and Adam Goucher and Lukas Gross and Katia Gil Guzman and John Hallman and Jackie Hehir and Johannes Heidecke and Alec Helyar and Haitang Hu and Romain Huet and Jacob Huh and Saachi Jain and Zach Johnson and Chris Koch and Irina Kofman and Dominik Kundel and Jason Kwon and Volodymyr Kyrylov and Elaine Ya Le and Guillaume Leclerc and James Park Lennon and Scott Lessans and Mario Lezcano-Casado and Yuanzhi Li and Zhuohan Li and Ji Lin and Jordan Liss and Lily and Liu and Jiancheng Liu and Kevin Lu and Chris Lu and Zoran Martinovic and Lindsay McCallum and Josh McGrath and Scott McKinney and Aidan McLaughlin and Song Mei and Steve Mostovoy and Tong Mu and Gideon Myles and Alexander Neitz and Alex Nichol and Jakub Pachocki and Alex Paino and Dana Palmie and Ashley Pantuliano and Giambattista Parascandolo and Jongsoo Park and Leher Pathak and Carolina Paz and Ludovic Peran and Dmitry Pimenov and Michelle Pokrass and Elizabeth Proehl and Huida Qiu and Gaby Raila and Filippo Raso and Hongyu Ren and Kimmy Richardson and David Robinson and Bob Rotsted and Hadi Salman and Suvansh Sanjeev and Max Schwarzer and D. Sculley and Harshit Sikchi and Kendal Simon and Karan Singhal and Yang Song and Dane Stuckey and Zhiqing Sun and Philippe Tillet and Sam Toizer and Foivos Tsimpourlas and Nikhil Vyas and Eric Wallace and Xin Wang and Miles Wang and Olivia Watkins and Kevin Weil and Amy Wendling and Kevin Whinnery and Cedric Whitney and Hannah Wong and Lin Yang and Yu Yang and Michihiro Yasunaga and Kristen Ying and Wojciech Zaremba and Wenting Zhan and Cyril Zhang and Brian Zhang and Eddie Zhang and Shengjia Zhao},
      year={2025},
      eprint={2508.10925},
      archivePrefix={arXiv},
      primaryClass={cs.CL},
      url={https://arxiv.org/abs/2508.10925}, 
}

@misc{deepseekai2025deepseekv3technicalreport,
      title={DeepSeek-V3 Technical Report}, 
      author={DeepSeek-AI and Aixin Liu and Bei Feng and Bing Xue and Bingxuan Wang and Bochao Wu and Chengda Lu and Chenggang Zhao and Chengqi Deng and Chenyu Zhang and Chong Ruan and Damai Dai and Daya Guo and Dejian Yang and Deli Chen and Dongjie Ji and Erhang Li and Fangyun Lin and Fucong Dai and Fuli Luo and Guangbo Hao and Guanting Chen and Guowei Li and H. Zhang and Han Bao and Hanwei Xu and Haocheng Wang and Haowei Zhang and Honghui Ding and Huajian Xin and Huazuo Gao and Hui Li and Hui Qu and J. L. Cai and Jian Liang and Jianzhong Guo and Jiaqi Ni and Jiashi Li and Jiawei Wang and Jin Chen and Jingchang Chen and Jingyang Yuan and Junjie Qiu and Junlong Li and Junxiao Song and Kai Dong and Kai Hu and Kaige Gao and Kang Guan and Kexin Huang and Kuai Yu and Lean Wang and Lecong Zhang and Lei Xu and Leyi Xia and Liang Zhao and Litong Wang and Liyue Zhang and Meng Li and Miaojun Wang and Mingchuan Zhang and Minghua Zhang and Minghui Tang and Mingming Li and Ning Tian and Panpan Huang and Peiyi Wang and Peng Zhang and Qiancheng Wang and Qihao Zhu and Qinyu Chen and Qiushi Du and R. J. Chen and R. L. Jin and Ruiqi Ge and Ruisong Zhang and Ruizhe Pan and Runji Wang and Runxin Xu and Ruoyu Zhang and Ruyi Chen and S. S. Li and Shanghao Lu and Shangyan Zhou and Shanhuang Chen and Shaoqing Wu and Shengfeng Ye and Shengfeng Ye and Shirong Ma and Shiyu Wang and Shuang Zhou and Shuiping Yu and Shunfeng Zhou and Shuting Pan and T. Wang and Tao Yun and Tian Pei and Tianyu Sun and W. L. Xiao and Wangding Zeng and Wanjia Zhao and Wei An and Wen Liu and Wenfeng Liang and Wenjun Gao and Wenqin Yu and Wentao Zhang and X. Q. Li and Xiangyue Jin and Xianzu Wang and Xiao Bi and Xiaodong Liu and Xiaohan Wang and Xiaojin Shen and Xiaokang Chen and Xiaokang Zhang and Xiaosha Chen and Xiaotao Nie and Xiaowen Sun and Xiaoxiang Wang and Xin Cheng and Xin Liu and Xin Xie and Xingchao Liu and Xingkai Yu and Xinnan Song and Xinxia Shan and Xinyi Zhou and Xinyu Yang and Xinyuan Li and Xuecheng Su and Xuheng Lin and Y. K. Li and Y. Q. Wang and Y. X. Wei and Y. X. Zhu and Yang Zhang and Yanhong Xu and Yanhong Xu and Yanping Huang and Yao Li and Yao Zhao and Yaofeng Sun and Yaohui Li and Yaohui Wang and Yi Yu and Yi Zheng and Yichao Zhang and Yifan Shi and Yiliang Xiong and Ying He and Ying Tang and Yishi Piao and Yisong Wang and Yixuan Tan and Yiyang Ma and Yiyuan Liu and Yongqiang Guo and Yu Wu and Yuan Ou and Yuchen Zhu and Yuduan Wang and Yue Gong and Yuheng Zou and Yujia He and Yukun Zha and Yunfan Xiong and Yunxian Ma and Yuting Yan and Yuxiang Luo and Yuxiang You and Yuxuan Liu and Yuyang Zhou and Z. F. Wu and Z. Z. Ren and Zehui Ren and Zhangli Sha and Zhe Fu and Zhean Xu and Zhen Huang and Zhen Zhang and Zhenda Xie and Zhengyan Zhang and Zhewen Hao and Zhibin Gou and Zhicheng Ma and Zhigang Yan and Zhihong Shao and Zhipeng Xu and Zhiyu Wu and Zhongyu Zhang and Zhuoshu Li and Zihui Gu and Zijia Zhu and Zijun Liu and Zilin Li and Ziwei Xie and Ziyang Song and Ziyi Gao and Zizheng Pan},
      year={2025},
      eprint={2412.19437},
      archivePrefix={arXiv},
      primaryClass={cs.CL},
      url={https://arxiv.org/abs/2412.19437}, 
}

@misc{yang2025qwen3technicalreport,
      title={Qwen3 Technical Report}, 
      author={An Yang and Anfeng Li and Baosong Yang and Beichen Zhang and Binyuan Hui and Bo Zheng and Bowen Yu and Chang Gao and Chengen Huang and Chenxu Lv and Chujie Zheng and Dayiheng Liu and Fan Zhou and Fei Huang and Feng Hu and Hao Ge and Haoran Wei and Huan Lin and Jialong Tang and Jian Yang and Jianhong Tu and Jianwei Zhang and Jianxin Yang and Jiaxi Yang and Jing Zhou and Jingren Zhou and Junyang Lin and Kai Dang and Keqin Bao and Kexin Yang and Le Yu and Lianghao Deng and Mei Li and Mingfeng Xue and Mingze Li and Pei Zhang and Peng Wang and Qin Zhu and Rui Men and Ruize Gao and Shixuan Liu and Shuang Luo and Tianhao Li and Tianyi Tang and Wenbiao Yin and Xingzhang Ren and Xinyu Wang and Xinyu Zhang and Xuancheng Ren and Yang Fan and Yang Su and Yichang Zhang and Yinger Zhang and Yu Wan and Yuqiong Liu and Zekun Wang and Zeyu Cui and Zhenru Zhang and Zhipeng Zhou and Zihan Qiu},
      year={2025},
      eprint={2505.09388},
      archivePrefix={arXiv},
      primaryClass={cs.CL},
      url={https://arxiv.org/abs/2505.09388}, 
}

@misc{kimiteam2025kimik2openagentic,
      title={Kimi K2: Open Agentic Intelligence}, 
      author={Kimi Team and Yifan Bai and Yiping Bao and Guanduo Chen and Jiahao Chen and Ningxin Chen and Ruijue Chen and Yanru Chen and Yuankun Chen and Yutian Chen and Zhuofu Chen and Jialei Cui and Hao Ding and Mengnan Dong and Angang Du and Chenzhuang Du and Dikang Du and Yulun Du and Yu Fan and Yichen Feng and Kelin Fu and Bofei Gao and Hongcheng Gao and Peizhong Gao and Tong Gao and Xinran Gu and Longyu Guan and Haiqing Guo and Jianhang Guo and Hao Hu and Xiaoru Hao and Tianhong He and Weiran He and Wenyang He and Chao Hong and Yangyang Hu and Zhenxing Hu and Weixiao Huang and Zhiqi Huang and Zihao Huang and Tao Jiang and Zhejun Jiang and Xinyi Jin and Yongsheng Kang and Guokun Lai and Cheng Li and Fang Li and Haoyang Li and Ming Li and Wentao Li and Yanhao Li and Yiwei Li and Zhaowei Li and Zheming Li and Hongzhan Lin and Xiaohan Lin and Zongyu Lin and Chengyin Liu and Chenyu Liu and Hongzhang Liu and Jingyuan Liu and Junqi Liu and Liang Liu and Shaowei Liu and T. Y. Liu and Tianwei Liu and Weizhou Liu and Yangyang Liu and Yibo Liu and Yiping Liu and Yue Liu and Zhengying Liu and Enzhe Lu and Lijun Lu and Shengling Ma and Xinyu Ma and Yingwei Ma and Shaoguang Mao and Jie Mei and Xin Men and Yibo Miao and Siyuan Pan and Yebo Peng and Ruoyu Qin and Bowen Qu and Zeyu Shang and Lidong Shi and Shengyuan Shi and Feifan Song and Jianlin Su and Zhengyuan Su and Xinjie Sun and Flood Sung and Heyi Tang and Jiawen Tao and Qifeng Teng and Chensi Wang and Dinglu Wang and Feng Wang and Haiming Wang and Jianzhou Wang and Jiaxing Wang and Jinhong Wang and Shengjie Wang and Shuyi Wang and Yao Wang and Yejie Wang and Yiqin Wang and Yuxin Wang and Yuzhi Wang and Zhaoji Wang and Zhengtao Wang and Zhexu Wang and Chu Wei and Qianqian Wei and Wenhao Wu and Xingzhe Wu and Yuxin Wu and Chenjun Xiao and Xiaotong Xie and Weimin Xiong and Boyu Xu and Jing Xu and Jinjing Xu and L. H. Xu and Lin Xu and Suting Xu and Weixin Xu and Xinran Xu and Yangchuan Xu and Ziyao Xu and Junjie Yan and Yuzi Yan and Xiaofei Yang and Ying Yang and Zhen Yang and Zhilin Yang and Zonghan Yang and Haotian Yao and Xingcheng Yao and Wenjie Ye and Zhuorui Ye and Bohong Yin and Longhui Yu and Enming Yuan and Hongbang Yuan and Mengjie Yuan and Haobing Zhan and Dehao Zhang and Hao Zhang and Wanlu Zhang and Xiaobin Zhang and Yangkun Zhang and Yizhi Zhang and Yongting Zhang and Yu Zhang and Yutao Zhang and Yutong Zhang and Zheng Zhang and Haotian Zhao and Yikai Zhao and Huabin Zheng and Shaojie Zheng and Jianren Zhou and Xinyu Zhou and Zaida Zhou and Zhen Zhu and Weiyu Zhuang and Xinxing Zu},
      year={2025},
      eprint={2507.20534},
      archivePrefix={arXiv},
      primaryClass={cs.LG},
      url={https://arxiv.org/abs/2507.20534}, 
}

@misc{shazeer2017outrageouslylargeneuralnetworks,
      title={Outrageously Large Neural Networks: The Sparsely-Gated Mixture-of-Experts Layer}, 
      author={Noam Shazeer and Azalia Mirhoseini and Krzysztof Maziarz and Andy Davis and Quoc Le and Geoffrey Hinton and Jeff Dean},
      year={2017},
      eprint={1701.06538},
      archivePrefix={arXiv},
      primaryClass={cs.LG},
      url={https://arxiv.org/abs/1701.06538}, 
}

@misc{rozière2024codellamaopenfoundation,
      title={Code Llama: Open Foundation Models for Code}, 
      author={Baptiste Rozière and Jonas Gehring and Fabian Gloeckle and Sten Sootla and Itai Gat and Xiaoqing Ellen Tan and Yossi Adi and Jingyu Liu and Romain Sauvestre and Tal Remez and Jérémy Rapin and Artyom Kozhevnikov and Ivan Evtimov and Joanna Bitton and Manish Bhatt and Cristian Canton Ferrer and Aaron Grattafiori and Wenhan Xiong and Alexandre Défossez and Jade Copet and Faisal Azhar and Hugo Touvron and Louis Martin and Nicolas Usunier and Thomas Scialom and Gabriel Synnaeve},
      year={2024},
      eprint={2308.12950},
      archivePrefix={arXiv},
      primaryClass={cs.CL},
      url={https://arxiv.org/abs/2308.12950}, 
}

@article{Luo_2022,
   title={BioGPT: generative pre-trained transformer for biomedical text generation and mining},
   volume={23},
   ISSN={1477-4054},
   url={http://dx.doi.org/10.1093/bib/bbac409},
   DOI={10.1093/bib/bbac409},
   number={6},
   journal={Briefings in Bioinformatics},
   publisher={Oxford University Press (OUP)},
   author={Luo, Renqian and Sun, Liai and Xia, Yingce and Qin, Tao and Zhang, Sheng and Poon, Hoifung and Liu, Tie-Yan},
   year={2022},
   month=sep
}

@misc{hui2024qwen25codertechnicalreport,
      title={Qwen2.5-Coder Technical Report}, 
      author={Binyuan Hui and Jian Yang and Zeyu Cui and Jiaxi Yang and Dayiheng Liu and Lei Zhang and Tianyu Liu and Jiajun Zhang and Bowen Yu and Keming Lu and Kai Dang and Yang Fan and Yichang Zhang and An Yang and Rui Men and Fei Huang and Bo Zheng and Yibo Miao and Shanghaoran Quan and Yunlong Feng and Xingzhang Ren and Xuancheng Ren and Jingren Zhou and Junyang Lin},
      year={2024},
      eprint={2409.12186},
      archivePrefix={arXiv},
      primaryClass={cs.CL},
      url={https://arxiv.org/abs/2409.12186}, 
}

@misc{wu2023bloomberggptlargelanguagemodel,
      title={BloombergGPT: A Large Language Model for Finance}, 
      author={Shijie Wu and Ozan Irsoy and Steven Lu and Vadim Dabravolski and Mark Dredze and Sebastian Gehrmann and Prabhanjan Kambadur and David Rosenberg and Gideon Mann},
      year={2023},
      eprint={2303.17564},
      archivePrefix={arXiv},
      primaryClass={cs.LG},
      url={https://arxiv.org/abs/2303.17564}, 
}

@misc{yang2023fingptopensourcefinanciallarge,
      title={FinGPT: Open-Source Financial Large Language Models}, 
      author={Hongyang Yang and Xiao-Yang Liu and Christina Dan Wang},
      year={2023},
      eprint={2306.06031},
      archivePrefix={arXiv},
      primaryClass={q-fin.ST},
      url={https://arxiv.org/abs/2306.06031}, 
}

@misc{cottier2025risingcoststrainingfrontier,
      title={The rising costs of training frontier AI models}, 
      author={Ben Cottier and Robi Rahman and Loredana Fattorini and Nestor Maslej and Tamay Besiroglu and David Owen},
      year={2025},
      eprint={2405.21015},
      archivePrefix={arXiv},
      primaryClass={cs.CY},
      url={https://arxiv.org/abs/2405.21015}, 
}

@misc{grattafiori2024llama3herdmodels,
      title={The Llama 3 Herd of Models}, 
      author={Aaron Grattafiori and Abhimanyu Dubey and Abhinav Jauhri and Abhinav Pandey and Abhishek Kadian and Ahmad Al-Dahle and Aiesha Letman and Akhil Mathur and Alan Schelten and Alex Vaughan and Amy Yang and Angela Fan and Anirudh Goyal and Anthony Hartshorn and Aobo Yang and Archi Mitra and Archie Sravankumar and Artem Korenev and Arthur Hinsvark and Arun Rao and Aston Zhang and Aurelien Rodriguez and Austen Gregerson and Ava Spataru and Baptiste Roziere and Bethany Biron and Binh Tang and Bobbie Chern and Charlotte Caucheteux and Chaya Nayak and Chloe Bi and Chris Marra and Chris McConnell and Christian Keller and Christophe Touret and Chunyang Wu and Corinne Wong and Cristian Canton Ferrer and Cyrus Nikolaidis and Damien Allonsius and Daniel Song and Danielle Pintz and Danny Livshits and Danny Wyatt and David Esiobu and Dhruv Choudhary and Dhruv Mahajan and Diego Garcia-Olano and Diego Perino and Dieuwke Hupkes and Egor Lakomkin and Ehab AlBadawy and Elina Lobanova and Emily Dinan and Eric Michael Smith and Filip Radenovic and Francisco Guzmán and Frank Zhang and Gabriel Synnaeve and Gabrielle Lee and Georgia Lewis Anderson and Govind Thattai and Graeme Nail and Gregoire Mialon and Guan Pang and Guillem Cucurell and Hailey Nguyen and Hannah Korevaar and Hu Xu and Hugo Touvron and Iliyan Zarov and Imanol Arrieta Ibarra and Isabel Kloumann and Ishan Misra and Ivan Evtimov and Jack Zhang and Jade Copet and Jaewon Lee and Jan Geffert and Jana Vranes and Jason Park and Jay Mahadeokar and Jeet Shah and Jelmer van der Linde and Jennifer Billock and Jenny Hong and Jenya Lee and Jeremy Fu and Jianfeng Chi and Jianyu Huang and Jiawen Liu and Jie Wang and Jiecao Yu and Joanna Bitton and Joe Spisak and Jongsoo Park and Joseph Rocca and Joshua Johnstun and Joshua Saxe and Junteng Jia and Kalyan Vasuden Alwala and Karthik Prasad and Kartikeya Upasani and Kate Plawiak and Ke Li and Kenneth Heafield and Kevin Stone and Khalid El-Arini and Krithika Iyer and Kshitiz Malik and Kuenley Chiu and Kunal Bhalla and Kushal Lakhotia and Lauren Rantala-Yeary and Laurens van der Maaten and Lawrence Chen and Liang Tan and Liz Jenkins and Louis Martin and Lovish Madaan and Lubo Malo and Lukas Blecher and Lukas Landzaat and Luke de Oliveira and Madeline Muzzi and Mahesh Pasupuleti and Mannat Singh and Manohar Paluri and Marcin Kardas and Maria Tsimpoukelli and Mathew Oldham and Mathieu Rita and Maya Pavlova and Melanie Kambadur and Mike Lewis and Min Si and Mitesh Kumar Singh and Mona Hassan and Naman Goyal and Narjes Torabi and Nikolay Bashlykov and Nikolay Bogoychev and Niladri Chatterji and Ning Zhang and Olivier Duchenne and Onur Çelebi and Patrick Alrassy and Pengchuan Zhang and Pengwei Li and Petar Vasic and Peter Weng and Prajjwal Bhargava and Pratik Dubal and Praveen Krishnan and Punit Singh Koura and Puxin Xu and Qing He and Qingxiao Dong and Ragavan Srinivasan and Raj Ganapathy and Ramon Calderer and Ricardo Silveira Cabral and Robert Stojnic and Roberta Raileanu and Rohan Maheswari and Rohit Girdhar and Rohit Patel and Romain Sauvestre and Ronnie Polidoro and Roshan Sumbaly and Ross Taylor and Ruan Silva and Rui Hou and Rui Wang and Saghar Hosseini and Sahana Chennabasappa and Sanjay Singh and Sean Bell and Seohyun Sonia Kim and Sergey Edunov and Shaoliang Nie and Sharan Narang and Sharath Raparthy and Sheng Shen and Shengye Wan and Shruti Bhosale and Shun Zhang and Simon Vandenhende and Soumya Batra and Spencer Whitman and Sten Sootla and Stephane Collot and Suchin Gururangan and Sydney Borodinsky and Tamar Herman and Tara Fowler and Tarek Sheasha and Thomas Georgiou and Thomas Scialom and Tobias Speckbacher and Todor Mihaylov and Tong Xiao and Ujjwal Karn and Vedanuj Goswami and Vibhor Gupta and Vignesh Ramanathan and Viktor Kerkez and Vincent Gonguet and Virginie Do and Vish Vogeti and Vítor Albiero and Vladan Petrovic and Weiwei Chu and Wenhan Xiong and Wenyin Fu and Whitney Meers and Xavier Martinet and Xiaodong Wang and Xiaofang Wang and Xiaoqing Ellen Tan and Xide Xia and Xinfeng Xie and Xuchao Jia and Xuewei Wang and Yaelle Goldschlag and Yashesh Gaur and Yasmine Babaei and Yi Wen and Yiwen Song and Yuchen Zhang and Yue Li and Yuning Mao and Zacharie Delpierre Coudert and Zheng Yan and Zhengxing Chen and Zoe Papakipos and Aaditya Singh and Aayushi Srivastava and Abha Jain and Adam Kelsey and Adam Shajnfeld and Adithya Gangidi and Adolfo Victoria and Ahuva Goldstand and Ajay Menon and Ajay Sharma and Alex Boesenberg and Alexei Baevski and Allie Feinstein and Amanda Kallet and Amit Sangani and Amos Teo and Anam Yunus and Andrei Lupu and Andres Alvarado and Andrew Caples and Andrew Gu and Andrew Ho and Andrew Poulton and Andrew Ryan and Ankit Ramchandani and Annie Dong and Annie Franco and Anuj Goyal and Aparajita Saraf and Arkabandhu Chowdhury and Ashley Gabriel and Ashwin Bharambe and Assaf Eisenman and Azadeh Yazdan and Beau James and Ben Maurer and Benjamin Leonhardi and Bernie Huang and Beth Loyd and Beto De Paola and Bhargavi Paranjape and Bing Liu and Bo Wu and Boyu Ni and Braden Hancock and Bram Wasti and Brandon Spence and Brani Stojkovic and Brian Gamido and Britt Montalvo and Carl Parker and Carly Burton and Catalina Mejia and Ce Liu and Changhan Wang and Changkyu Kim and Chao Zhou and Chester Hu and Ching-Hsiang Chu and Chris Cai and Chris Tindal and Christoph Feichtenhofer and Cynthia Gao and Damon Civin and Dana Beaty and Daniel Kreymer and Daniel Li and David Adkins and David Xu and Davide Testuggine and Delia David and Devi Parikh and Diana Liskovich and Didem Foss and Dingkang Wang and Duc Le and Dustin Holland and Edward Dowling and Eissa Jamil and Elaine Montgomery and Eleonora Presani and Emily Hahn and Emily Wood and Eric-Tuan Le and Erik Brinkman and Esteban Arcaute and Evan Dunbar and Evan Smothers and Fei Sun and Felix Kreuk and Feng Tian and Filippos Kokkinos and Firat Ozgenel and Francesco Caggioni and Frank Kanayet and Frank Seide and Gabriela Medina Florez and Gabriella Schwarz and Gada Badeer and Georgia Swee and Gil Halpern and Grant Herman and Grigory Sizov and Guangyi and Zhang and Guna Lakshminarayanan and Hakan Inan and Hamid Shojanazeri and Han Zou and Hannah Wang and Hanwen Zha and Haroun Habeeb and Harrison Rudolph and Helen Suk and Henry Aspegren and Hunter Goldman and Hongyuan Zhan and Ibrahim Damlaj and Igor Molybog and Igor Tufanov and Ilias Leontiadis and Irina-Elena Veliche and Itai Gat and Jake Weissman and James Geboski and James Kohli and Janice Lam and Japhet Asher and Jean-Baptiste Gaya and Jeff Marcus and Jeff Tang and Jennifer Chan and Jenny Zhen and Jeremy Reizenstein and Jeremy Teboul and Jessica Zhong and Jian Jin and Jingyi Yang and Joe Cummings and Jon Carvill and Jon Shepard and Jonathan McPhie and Jonathan Torres and Josh Ginsburg and Junjie Wang and Kai Wu and Kam Hou U and Karan Saxena and Kartikay Khandelwal and Katayoun Zand and Kathy Matosich and Kaushik Veeraraghavan and Kelly Michelena and Keqian Li and Kiran Jagadeesh and Kun Huang and Kunal Chawla and Kyle Huang and Lailin Chen and Lakshya Garg and Lavender A and Leandro Silva and Lee Bell and Lei Zhang and Liangpeng Guo and Licheng Yu and Liron Moshkovich and Luca Wehrstedt and Madian Khabsa and Manav Avalani and Manish Bhatt and Martynas Mankus and Matan Hasson and Matthew Lennie and Matthias Reso and Maxim Groshev and Maxim Naumov and Maya Lathi and Meghan Keneally and Miao Liu and Michael L. Seltzer and Michal Valko and Michelle Restrepo and Mihir Patel and Mik Vyatskov and Mikayel Samvelyan and Mike Clark and Mike Macey and Mike Wang and Miquel Jubert Hermoso and Mo Metanat and Mohammad Rastegari and Munish Bansal and Nandhini Santhanam and Natascha Parks and Natasha White and Navyata Bawa and Nayan Singhal and Nick Egebo and Nicolas Usunier and Nikhil Mehta and Nikolay Pavlovich Laptev and Ning Dong and Norman Cheng and Oleg Chernoguz and Olivia Hart and Omkar Salpekar and Ozlem Kalinli and Parkin Kent and Parth Parekh and Paul Saab and Pavan Balaji and Pedro Rittner and Philip Bontrager and Pierre Roux and Piotr Dollar and Polina Zvyagina and Prashant Ratanchandani and Pritish Yuvraj and Qian Liang and Rachad Alao and Rachel Rodriguez and Rafi Ayub and Raghotham Murthy and Raghu Nayani and Rahul Mitra and Rangaprabhu Parthasarathy and Raymond Li and Rebekkah Hogan and Robin Battey and Rocky Wang and Russ Howes and Ruty Rinott and Sachin Mehta and Sachin Siby and Sai Jayesh Bondu and Samyak Datta and Sara Chugh and Sara Hunt and Sargun Dhillon and Sasha Sidorov and Satadru Pan and Saurabh Mahajan and Saurabh Verma and Seiji Yamamoto and Sharadh Ramaswamy and Shaun Lindsay and Shaun Lindsay and Sheng Feng and Shenghao Lin and Shengxin Cindy Zha and Shishir Patil and Shiva Shankar and Shuqiang Zhang and Shuqiang Zhang and Sinong Wang and Sneha Agarwal and Soji Sajuyigbe and Soumith Chintala and Stephanie Max and Stephen Chen and Steve Kehoe and Steve Satterfield and Sudarshan Govindaprasad and Sumit Gupta and Summer Deng and Sungmin Cho and Sunny Virk and Suraj Subramanian and Sy Choudhury and Sydney Goldman and Tal Remez and Tamar Glaser and Tamara Best and Thilo Koehler and Thomas Robinson and Tianhe Li and Tianjun Zhang and Tim Matthews and Timothy Chou and Tzook Shaked and Varun Vontimitta and Victoria Ajayi and Victoria Montanez and Vijai Mohan and Vinay Satish Kumar and Vishal Mangla and Vlad Ionescu and Vlad Poenaru and Vlad Tiberiu Mihailescu and Vladimir Ivanov and Wei Li and Wenchen Wang and Wenwen Jiang and Wes Bouaziz and Will Constable and Xiaocheng Tang and Xiaojian Wu and Xiaolan Wang and Xilun Wu and Xinbo Gao and Yaniv Kleinman and Yanjun Chen and Ye Hu and Ye Jia and Ye Qi and Yenda Li and Yilin Zhang and Ying Zhang and Yossi Adi and Youngjin Nam and Yu and Wang and Yu Zhao and Yuchen Hao and Yundi Qian and Yunlu Li and Yuzi He and Zach Rait and Zachary DeVito and Zef Rosnbrick and Zhaoduo Wen and Zhenyu Yang and Zhiwei Zhao and Zhiyu Ma},
      year={2024},
      eprint={2407.21783},
      archivePrefix={arXiv},
      primaryClass={cs.AI},
      url={https://arxiv.org/abs/2407.21783}, 
}

@article{Jacobs1991Adaptive,
  author    = {Jacobs, Robert A. and Jordan, Michael I. and Nowlan, Steven J. and Hinton, Geoffrey E.},
  title     = {Adaptive Mixtures of Local Experts},
  journal   = {Neural Computation},
  volume    = {3},
  number    = {1},
  pages     = {79--87},
  year      = {1991},
  doi       = {10.1162/neco.1991.3.1.79}
}

@inproceedings{Casta_o_2023,
   title={Exploring the Carbon Footprint of Hugging Face’s ML Models: A Repository Mining Study},
   url={http://dx.doi.org/10.1109/ESEM56168.2023.10304801},
   DOI={10.1109/esem56168.2023.10304801},
   booktitle={2023 ACM/IEEE International Symposium on Empirical Software Engineering and Measurement (ESEM)},
   publisher={IEEE},
   author={Castaño, Joel and Martínez-Fernández, Silverio and Franch, Xavier and Bogner, Justus},
   year={2023},
   month=oct, pages={1–12}
}

@article{survery_llm_powers,
author = {Yang, Jingfeng and Jin, Hongye and Tang, Ruixiang and Han, Xiaotian and Feng, Qizhang and Jiang, Haoming and Zhong, Shaochen and Yin, Bing and Hu, Xia},
title = {Harnessing the Power of LLMs in Practice: A Survey on ChatGPT and Beyond},
year = {2024},
issue_date = {July 2024},
publisher = {Association for Computing Machinery},
address = {New York, NY, USA},
volume = {18},
number = {6},
issn = {1556-4681},
url = {https://doi.org/10.1145/3649506},
doi = {10.1145/3649506},
journal = {ACM Trans. Knowl. Discov. Data},
month = apr,
articleno = {160},
numpages = {32}
}

@misc{kaplan2020scalinglawsneurallanguage,
      title={Scaling Laws for Neural Language Models}, 
      author={Jared Kaplan and Sam McCandlish and Tom Henighan and Tom B. Brown and Benjamin Chess and Rewon Child and Scott Gray and Alec Radford and Jeffrey Wu and Dario Amodei},
      year={2020},
      eprint={2001.08361},
      archivePrefix={arXiv},
      primaryClass={cs.LG},
      url={https://arxiv.org/abs/2001.08361}, 
}

@misc{haque2025catastrophicforgettingllmscomparative,
      title={Catastrophic Forgetting in LLMs: A Comparative Analysis Across Language Tasks}, 
      author={Naimul Haque},
      year={2025},
      eprint={2504.01241},
      archivePrefix={arXiv},
      primaryClass={cs.CL},
      url={https://arxiv.org/abs/2504.01241}, 
}

@ARTICLE{cf_luo,
  author={Luo, Yun and Yang, Zhen and Meng, Fandong and Li, Yafu and Zhou, Jie and Zhang, Yue},
  journal={IEEE Transactions on Audio, Speech and Language Processing}, 
  title={An Empirical Study of Catastrophic Forgetting in Large Language Models During Continual Fine-Tuning}, 
  year={2025},
  volume={33},
  number={},
  pages={3776-3786},
  doi={10.1109/TASLPRO.2025.3606231}
}

@article{Osborne_2024,
   title={The AI community building the future? A quantitative analysis of development activity on Hugging Face Hub},
   volume={7},
   ISSN={2432-2725},
   url={http://dx.doi.org/10.1007/s42001-024-00300-8},
   DOI={10.1007/s42001-024-00300-8},
   number={2},
   journal={Journal of Computational Social Science},
   publisher={Springer Science and Business Media LLC},
   author={Osborne, Cailean and Ding, Jennifer and Kirk, Hannah Rose},
   year={2024},
   month=jun, pages={2067–2105} }

@misc{fedus2022switchtransformersscalingtrillion,
      title={Switch Transformers: Scaling to Trillion Parameter Models with Simple and Efficient Sparsity}, 
      author={William Fedus and Barret Zoph and Noam Shazeer},
      year={2022},
      eprint={2101.03961},
      archivePrefix={arXiv},
      primaryClass={cs.LG},
      url={https://arxiv.org/abs/2101.03961}, 
}

@misc{du2022glamefficientscalinglanguage,
      title={GLaM: Efficient Scaling of Language Models with Mixture-of-Experts}, 
      author={Nan Du and Yanping Huang and Andrew M. Dai and Simon Tong and Dmitry Lepikhin and Yuanzhong Xu and Maxim Krikun and Yanqi Zhou and Adams Wei Yu and Orhan Firat and Barret Zoph and Liam Fedus and Maarten Bosma and Zongwei Zhou and Tao Wang and Yu Emma Wang and Kellie Webster and Marie Pellat and Kevin Robinson and Kathleen Meier-Hellstern and Toju Duke and Lucas Dixon and Kun Zhang and Quoc V Le and Yonghui Wu and Zhifeng Chen and Claire Cui},
      year={2022},
      eprint={2112.06905},
      archivePrefix={arXiv},
      primaryClass={cs.CL},
      url={https://arxiv.org/abs/2112.06905}, 
}

@misc{mu2025comprehensivesurveymixtureofexpertsalgorithms,
      title={A Comprehensive Survey of Mixture-of-Experts: Algorithms, Theory, and Applications}, 
      author={Siyuan Mu and Sen Lin},
      year={2025},
      eprint={2503.07137},
      archivePrefix={arXiv},
      primaryClass={cs.LG},
      url={https://arxiv.org/abs/2503.07137}, 
}

@misc{li2022branchtrainmergeembarrassinglyparalleltraining,
      title={Branch-Train-Merge: Embarrassingly Parallel Training of Expert Language Models}, 
      author={Margaret Li and Suchin Gururangan and Tim Dettmers and Mike Lewis and Tim Althoff and Noah A. Smith and Luke Zettlemoyer},
      year={2022},
      eprint={2208.03306},
      archivePrefix={arXiv},
      primaryClass={cs.CL},
      url={https://arxiv.org/abs/2208.03306}, 
}

@misc{wortsman2022modelsoupsaveragingweights,
      title={Model soups: averaging weights of multiple fine-tuned models improves accuracy without increasing inference time}, 
      author={Mitchell Wortsman and Gabriel Ilharco and Samir Yitzhak Gadre and Rebecca Roelofs and Raphael Gontijo-Lopes and Ari S. Morcos and Hongseok Namkoong and Ali Farhadi and Yair Carmon and Simon Kornblith and Ludwig Schmidt},
      year={2022},
      eprint={2203.05482},
      archivePrefix={arXiv},
      primaryClass={cs.LG},
      url={https://arxiv.org/abs/2203.05482}, 
}

@misc{fu2024moamixturesparseattention,
      title={MoA: Mixture of Sparse Attention for Automatic Large Language Model Compression}, 
      author={Tianyu Fu and Haofeng Huang and Xuefei Ning and Genghan Zhang and Boju Chen and Tianqi Wu and Hongyi Wang and Zixiao Huang and Shiyao Li and Shengen Yan and Guohao Dai and Huazhong Yang and Yu Wang},
      year={2024},
      eprint={2406.14909},
      archivePrefix={arXiv},
      primaryClass={cs.LG},
      url={https://arxiv.org/abs/2406.14909}, 
}

@misc{chamma2023statisticallyvalidvariableimportance,
      title={Statistically Valid Variable Importance Assessment through Conditional Permutations}, 
      author={Ahmad Chamma and Denis A. Engemann and Bertrand Thirion},
      year={2023},
      eprint={2309.07593},
      archivePrefix={arXiv},
      primaryClass={cs.LG},
      url={https://arxiv.org/abs/2309.07593}, 
}

@ARTICLE{egyptian_dialect,
  author={Elnagar, Ashraf and Yagi, Sane M. and Nassif, Ali Bou and Shahin, Ismail and Salloum, Said A.},
  journal={IEEE Access}, 
  title={Systematic Literature Review of Dialectal Arabic: Identification and Detection}, 
  year={2021},
  volume={9},
  number={},
  pages={31010-31042},
  doi={10.1109/ACCESS.2021.3059504}}
